\documentclass[runningheads]{llncs}
\usepackage{graphicx}
\usepackage{multirow}
\usepackage{todonotes}
\usepackage{comment}
\usepackage{adjustbox}
\usepackage{amsmath}
\usepackage[utf8]{inputenc}
\usepackage[english]{babel}
\usepackage{color}
\usepackage{amsfonts}
\usepackage{url}
\usepackage[mathscr]{euscript}
 \let\mathscr\relax%
\usepackage[scr]{rsfso}
\usepackage{caption}
\usepackage[labelfont=bf]{caption}
\usepackage{tabularx,booktabs}
\newcolumntype{C}{>{\centering\arraybackslash}X} 
\usepackage{multicol}
\usepackage{algorithm2e}
\usepackage{booktabs,ragged2e}
\usepackage[utf8]{inputenc}
\usepackage{todonotes}
\usepackage{lineno}
\usepackage{makecell}
\usepackage[flushleft]{threeparttable}
\usepackage{subcaption}
\usepackage{color}

\usepackage[nolist,nohyperlinks]{acronym}
\usepackage{footnote}
\usepackage[perpage]{footmisc}
\usepackage[autostyle]{csquotes}
\makesavenoteenv{tabular}
\acrodef{DL}{Deep learning}
\acrodef{CNN}{Convolutional Neural Network}
\acrodef{ML}{Machine Learning}
\acrodef{mIoU}{mean Intersection over Union}
\acrodef{GI}{gastrointestinal}
\acrodef{AI}{Artificial Intelligence} 
\acrodef{CADx}{computer aided diagnosis} 
\acrodef{CRC}{Colorectal cancer}
\acrodef{DSC}{Dice Coefficient}
\acrodef{mDSC}{Dice Coefficient}
\acrodef{OOD}{Out-Of-Distribution}
\acrodef{SOTA}{State-of-the-art}
\acrodef{HD}{Hausdorff distance}
\acrodef{FANetv2}{Feedback Attention Network-v2}

\begin{document}


\title{Transformer-Enhanced Iterative Feedback Mechanism for Polyp Segmentation}

\titlerunning{Transformer-Enhanced Iterative Feedback Mechanism for Polyp Segmentation}
 \author{Nikhil Kumar Tomar\thanks{Corresponding author: Nikhil Kumar Tomar, Email: nikhilroxtomar@gmail.com}\inst{1} , Debesh Jha\inst{1} , Koushik Biswas\inst{1},  Tyler M. Berzin\inst{3} , Rajesh Keswani\inst{1} , Michael Wallace\inst{2} , Ulas Bagci\inst{1} }
 \institute{Machine \& Hybrid Intelligence Lab, Department of Radiology, Northwestern University, Chicago, USA \and Mayo Clinic, USA \and Beth Israel Deaconess Medical Center, Harvard Medical School, USA  }

  \maketitle      
\begin{abstract}
Colorectal cancer (CRC) is the third most common cause of cancer diagnosed in the United States and the second leading cause of cancer-related death among both genders. Notably, CRC is the leading cause of cancer in younger men less than 50 years old. Colonoscopy is considered the gold standard for the early diagnosis of CRC. Skills vary significantly among endoscopists, and a high miss rate is reported. Automated polyp segmentation can reduce the missed rates, and timely treatment is possible in the early stage. To address this challenge, we introduce \textit{\textbf{\ac{FANetv2}}}, an advanced encoder-decoder network designed to accurately segment polyps from colonoscopy images. Leveraging an initial input mask generated by Otsu thresholding, FANetv2 iteratively refines its binary segmentation masks through a novel feedback attention mechanism informed by the mask predictions of previous epochs. Additionally, it employs a text-guided approach that integrates essential information about the number (one or many) and size (small, medium, large) of polyps to further enhance its feature representation capabilities. This dual-task approach facilitates accurate polyp segmentation and aids in the auxiliary classification of polyp attributes, significantly boosting the model's performance. Our comprehensive evaluations on the publicly available BKAI-IGH and CVC-ClinicDB datasets demonstrate the superior performance of FANetv2, evidenced by high dice similarity coefficients (DSC) of 0.9186 and 0.9481, along with low Hausdorff distances of 2.83 and 3.19, respectively. The source code for FANetv2 is available at https://github.com/xxxxx/FANetv2.

\end{abstract}

\section{Introduction}
The Cancer statistic 2023~\cite{siegel2023cancer} estimates that CRC is the third leading cause of cancer-related incidence and second cancer-related death in the United States. There were around 153,020 new incidences of CRC and 52,550 deaths from colon and rectum cancer together. Although the overall rate of CRC has decreased due to the availability of advanced screening tests, the incidence rate for younger adults under 50 has increased by 1\%-2\%. Similarly, the overall mortality rate has improved due to the development of new treatment procedures, whereas the mortality rate under 55 years has risen significantly~\cite{siegel2023cancer}. However, some earlier studies have shown the lesion miss rate to be as high as 26\%~\cite{corley2014adenoma}. This highlights the demand for an accurate \ac{CADx} method for early detection of polyps and reducing the overall polyp miss rate.

Polyp segmentation is an active research area~\cite{jha2024transnetr,fan2020pranet,dong2021polyp,tomar2022fanet,tang2022duat,tomar2022tganet,zhang2021transfuse,zhao2021automatic}. Jha et al.~\cite{jha2024transnetr} proposed a novel deep learning architecture, TransNetR that utilizes the strength of transformer encoder and residual learning to obtain precise segmentation output masks even with out-of-distribution datasets that are entirely different than training set. Tomar et al.~\cite{tomar2022tganet} proposed a text-guided attention network (TGANet) to handle variations in polyp size (diminutive, regular, or large) and numbers of occurrence (one or many) for optimal polyp segmentation. Segment-anything models (SAM)~\cite{kirillov2023segment} have become popular recently. Since then, researchers have been exploring the utilization of SAM for medical images. For example, Ma et al.~\cite{ma2024segment} proposed MedSAM, a foundational model designed by fine-tuning the SAM model for different medical image segmentation tasks with large-scale medical image datasets, including polyps. Their finding concluded that MedSAM~\cite{ma2024segment} obtained better performance in terms of both qualitative and quantitative results compared to SAM~\cite{kirillov2023segment}, U-Net~\cite{ronneberger2015u}, and DeepLabv3+~\cite{chen2018encoder}.

Despite several advancements in the field, accurately identifying polyps with variable appearances remains an issue. Additionally, there is still potential for performance improvement to ensure a robust solution that could perform consistently across various lightning conditions and rare cases. To address this issue, we introduce novel \acf{FANetv2} architecture that utilizes the strength of iterative feedback attention to address the variability present in each image sample and text-guided attention to incorporate the information about the polyp occurrence and their sizes for improved polyp segmentation.  The primary contributions of this work are as follows:
\begin{enumerate}
    
    \item \textbf{Iterative feedback attention --- } We introduce a novel iterative feedback mechanism  which enhances the accuracy of polyp segmentation by leveraging the intrinsic variability of colonoscopy images. By iteratively refining segmentation masks using information from previous epochs, our method addresses the critical challenge of variable detection rates among endoscopists.

    \item \textbf{Text guided attention--- } FANetv2 incorporates text-guided attention that utilizes polyp characteristics such as a number of occurrences (one or many)  and size (small, medium, or large) as contextual cues for improving feature extraction. This integration of textual cues to the polyp images significantly improves segmentation accuracy. 

   \item \textbf{Dynamic masks refinement during testing ---} Using feedback information in the bridge and decoder part of the network helps refine the predicted masks in both the training and testing phases. During the testing phase, we iterate over the input image and update the input mask with the predicted mask {for up to three iterations (empirically set (see Supp. Table 1 \& Table 2))}. This dynamic refinement process boosts the segmentation performance in real time without requiring any post-processing steps. 

    \item \textbf{Comprehensive performance assessment --- } Experiments on BKAI-IGH~\cite{lan2021neounet} and CVC-ClinicDB~\cite{bernal2015wm} suggest that FANet exhibit superior capabilities by outperforming 12 \ac{SOTA} algorithms including several recent transformer based approach, which usually shows robust performance.
\end{enumerate}

\section{Method}
\begin{figure*} [!t]
    \centering
    \includegraphics[width=0.99\textwidth]{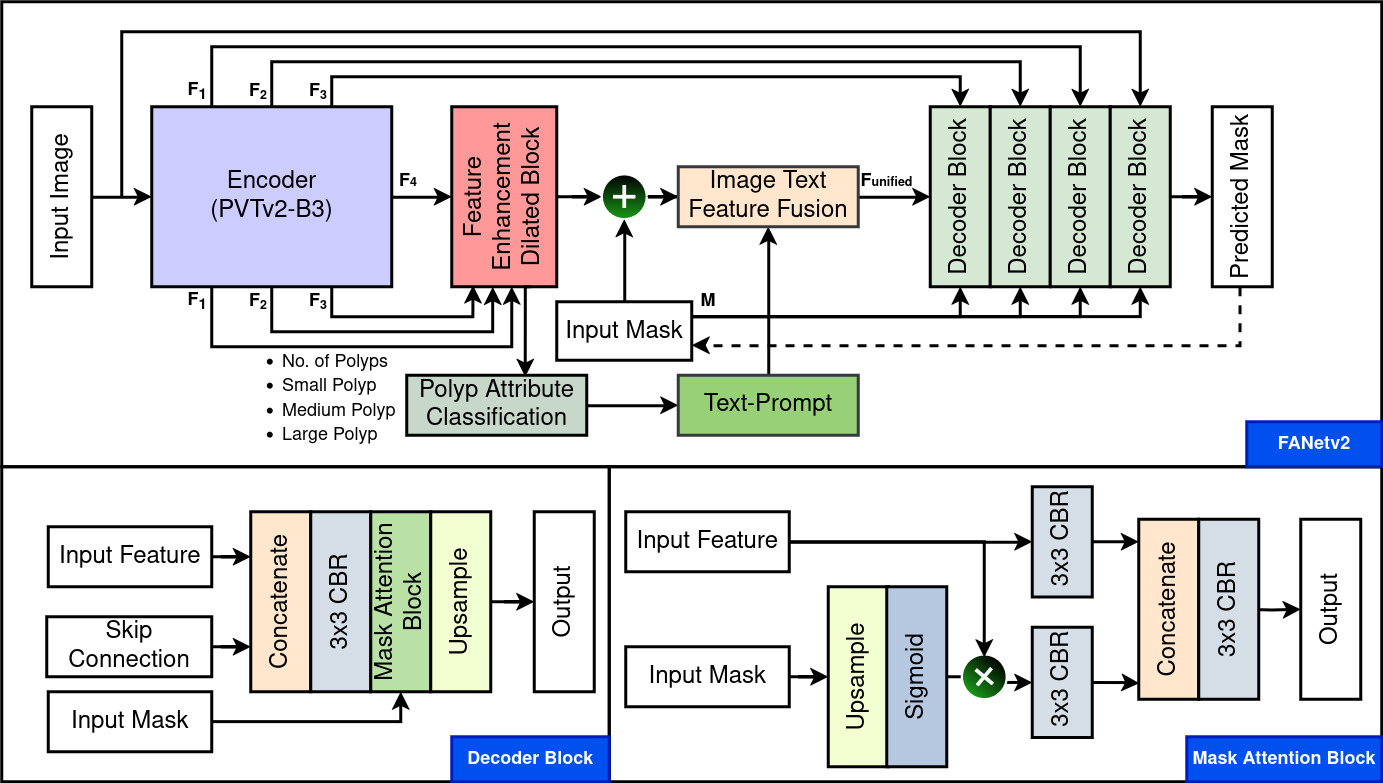}
\caption{Overview of the FANetv2 architecture. FANetv2  takes two inputs: \textit{a polyp image} and an \textit{initial input mask generated from Otsu thresholding}. The method is designed to perform two key tasks: \textit{an auxiliary polyp attribute classification} and the \textit{ polyp segmentation}. The input image is initially fed to the encoder, which forwards its output to the Feature Enhancement Block, whose output is used for \textit{polyp attribute classification} and the\textit{ rest of the network for polyp segmentation}. An innovative aspect of FANetv2 is that it uses an initial input mask and the polyp attributes to generate a unified feature representation, which is then passed to the decoder to predict the final segmentation mask. It is to be noted that FANetv2 has two key mechanisms: \textit{a feedback attention mechanism} that leverages input mask from the previous epoch to guide the proposed network to refine segmentation and a \textit{text-guided mechanism} that incorporates crucial information about the polyp, such as the number and size of polyps present within an image. These components work together to enhance the feature representation, thus improving the overall performance of the proposed FANetv2.}
    \label{fig:fanetv2}
    \vspace{-5mm}
\end{figure*}

FANetv2 is a specialized encoder-decoder architecture for polyp segmentation. It takes a polyp image and an initial input coarse-mask (generated using Otsu thresholding), then predicts a binary segmentation fine-mask. This predicted mask is used as the input mask for the next training epoch. FANetv2 performs two tasks: \textit{auxiliary polyp attribute classification} and \textit{main polyp segmentation}. The block diagram shown in Figure~\ref{fig:fanetv2} displays all components, including two key mechanisms: \textit{feedback attention}, using the previous epoch's predicted mask to guide the network, and a \textit{text-guided attention mechanism} incorporating important information about the number of polyps occurrence (one or many) and their sizes (small, medium, large). Together, these components enhance feature representation, improving FANetv2's overall performance.

\subsection{Pyramid Vision Transformer (PVT) Encoder}
The FANetv2 begins with a pre-trained encoder, where we have utilized a pyramid vision transformer (PVTv2). An input image $I \in \mathbb{R}^{H\times W\times 3}$ is fed to the PVT-encoder to obtain four distinct multi-scale feature maps $F_i \in \mathbb{R}^{\frac{H}{2^{i+1}} \times \frac{W}{2^{i+1}} \times C_i}$, where  $i \in \{1, 2, 3, 4\}$ and $C_i \in \{64, 128, 320, 512\}$. These feature maps are then used by the subsequent components of the FANetv2. 

\subsection{Feature Enhancement Dilated (FED) Block}
The Feature Enhancement Dilated (FED) block combines all four multi-scale feature maps from the PVT-encoder and feeds them to parallel dilated convolution layers to enhance the overall feature representation. The feature map $F_1$ is passed through a series of $3\times3$ Conv-BN-ReLU layers, which is then followed by $2\times2$ max-pooling to reduce its spatial dimension. The output of the max-pooling operation is then concatenated with feature map $F_2$ and then followed by a series of $3\times3$ Conv-BN-ReLU layers. The resulting feature map is again passed through a  $2\times2$ max-pooling operation. Like this, we concatenate remaining feature maps $F_3$ and $F_4$ and get a combined feature representation, which is then passed through a $1\times1$ Conv-BN-ReLU and three parallel convolution layer with a dilation rate of \{$6$, $12$, $18$\}. The output of the four convolution layers is concatenated and enhanced by a residual connection to deal with the vanishing gradient problem. The output of the FED block is utilized by both the auxiliary polyp attribute classification task and a main polyp segmentation task.

\subsection{Auxiliary Polyp Attribute Classification}
The output of FED block is passed through an adaptive average pooling layer to get a reduced representation. This reduced representation is used for four simple classification tasks, i.e., the number of polyps (one or many) and their size (small, medium, and large). Each individual size is predicted using a separate binary classification task. This is done as a single polyp image can contain multiple polyps, each of which can have a different size. The results from these classification tasks are then converted into a text prompt \textit{(e.g., a colorectal image with one small sized polyp, a colorectal image with many small and medium sized polyps or a colorectal image with many small sized polyps)}. This text prompt is then converted into an embedding by using sub-word byte-pair encoding~\cite{heinzerling2017bpemb}. The embeddings contain essential information regarding the number and size of the polyps present in the images, which is further combined with the image-level features from the FED block.  Over the training, the quality of the text prompt improves, which further helps to boost the overall performance of the FANetv2.

\subsection{Bridge -- Fusing Image and Textual Features}
Here, we integrate features from various components, i.e., FED block, input mask, and text-prompt embeddings, in order to have unified feature representation \{$F_{unified}$\}, which is further utilized by the decoder to generate the segmentation mask. The output of the FED block is added with the downsampled input mask and then followed by a series of $3\times3$ Conv-BN-ReLU layers. Next, we fuse the features with the text-prompt embeddings, incorporating crucial information regarding the number and size of polyps present within an image. By encoding these details, the network gains a deeper understanding of the specific characteristics and context of the polyps.

\subsection{Mask Attention  Block}
The proposed Mask Attention (MA) block is used in the decoder part of our FANetv2. It operates by performing an element-wise multiplication between the feature map and the input mask, which represents the segmentation mask predicted in the preceding epoch. This process enables the block to selectively emphasize regions of interest within the feature map, guided by the information encoded in the input mask. By focusing attention on relevant areas delineated by the mask, the Mask Attention block effectively refines segmentation boundaries and improves the overall accuracy of the segmentation process. 

The block begins with an element-wise multiplication between the feature map \{$F_{map}$\} and the resized input mask \{$M$\}, followed by $3\times3$ Conv-BN-ReLU layers. Next, the feature map \{$F_{map}$\} is passed to another set of $3\times3$ Conv-BN-ReLU layers and then followed by a concatenation \{$\cup$\}. The concatenated feature map is passed through a $1\times1$ Conv-BN-ReLU.

\begin{equation}
   \label{eq:mask_attention_equation}
   F_{output} = C_{1\times1}(C_{3\times3}(F_{map} \odot M) \cup C_{3\times3}(F_{map}))
\end{equation}

\subsection{Decoder}
The decoder in the proposed FANetv2 is composed of three different decoder blocks {$d_i \in 1, 2, 3$}, of which each takes the input feature map and concatenates it with the same scale feature map from the PVT-encoder. The concatenated feature map is then followed by two sets of $3\times3$ Conv-BN-ReLU layers, facilitating the learning of a semantic representation. Next, it is followed by the Mask Attention block, which provides spatial attention to correctly localize and highlight the features representing the polyps. The output of the Mask Attention block is then upsampled by a factor of two using bilinear interpolation. Similarly, follows the second and third decoder blocks, where we upsampled the feature map by a factor of two and four, respectively. The output from the third decoder block is followed by a concatenation with the input image, two sets of $3\times3$ Conv-BN-ReLU layers, and a Mask Attention block. The output is then passed to a $1\times1$ convolution layer with a sigmoid activation function to get the segmentation mask.

\section{Experiments and Results}
\subsection{Datasets:} We use BKAI-IGH~\cite{lan2021neounet} and CVC-ClinicDB~\cite{bernal2015wm} dataset for experimentation. The BKAI-IGH consists of 1000 images with the corresponding ground truth. The dataset was collected from two medical centers in Vietnam. It contains images captured using WLI, FICE, BLI and LCI. To train all the algorithms, we used 800 images for training, 100 for validation, and 100 for testing. Similarly, CVC-ClinicDB consists of 612 images collected from 31 video sequence. We have used 490 in the training, 61 in the validation, and 61 for the testing set.


\subsection{{Experimental Setup:}} We have trained all the models using the Pytorch~\cite{paszke2019pytorch} framework on the NVIDIA GeForce RTX 3090 GPU. All the images are first resized to $256 \times 256$ pixels. To improve the model's generalization capability and prevent overfitting,  we use basic data augmentation strategies, which include random rotation, vertical flipping, horizontal flipping, and coarse dropout. All models are trained on a similar hyperparameters configuration with a learning rate of $1e^{-4}$, batch size of 16, and an ADAM optimizer. We use a combination of binary cross-entropy and dice loss with equal weights as a loss function. In addition, we use an early stopping and \textit{ReduceLROnPlateau} to avoid overfitting.

\begin{table}[htpb!]
\footnotesize
\centering
\caption{Result of models trained and tested on BKAI-IGH~\cite{lan2021neounet}. }
 \begin{tabular} {l|c|c|c|c|c|c|c}
\toprule
\textbf{Method}  &\textbf{Publication} &\textbf{mDSC}  & \textbf{mIoU}  &\textbf{Recall}& \textbf{Precision} &\textbf{F2} &\textbf{HD} \\ 
\hline
U-Net~\cite{ronneberger2015u} &MICCAI 2015  &0.8286 &0.7599 &0.8295 &0.8999 &0.8264 &3.17\\
DeepLabV3+~\cite{chen2018encoder} &ECCV 2018 &0.8938 &0.8314 &0.8870 &0.9333 &0.8882 &2.90\\
\textcolor{blue}{FANet}~\cite{tomar2022fanet} &TNNLS 2021 &\underline{0.8205} &\underline{0.7570} &\underline{0.9012} &\underline{0.8300} &\underline{0.8404} &\underline{3.18}\\
PraNet~\cite{fan2020pranet} &MICCAI 2021 &0.8904 &0.8264 &0.8901 &0.9247 &0.8885 &2.94\\
Polyp-PVT~\cite{dong2021polyp}&ArXiv 2021 &0.8995 &0.8379 &0.9016 &0.9238 &0.8986 &2.88\\
UACANet~\cite{kim2021uacanet} &ACMMM 2021 &0.8945 &0.8275 &0.8870 &0.9297 &0.8882 &2.86	\\ 
DuAT~\cite{tang2022duat} &Arxiv 2022 &0.9140 &0.8563 &0.9038 &0.9437 &0.9066 &\textbf{2.77}\\
CaraNet~\cite{lou2022caranet}	&MIIP 2022 &0.8962 &0.8329 &0.8939 &0.9273 &0.8937 &2.91\\
LDNet~\cite{zhang2022lesion}&MICCAI 2022	&0.8927	&0.8254		&0.8867	&0.9153	&0.8874	&2.94 \\
TGANet~\cite{tomar2022tganet}&MICCAI 2022 &0.9023 &0.8409 &0.9025 &0.9208 &0.9002 &2.84\\
TransNetR~\cite{jha2024transnetr}& MIDL 2023 &0.9107 &0.8474 &0.8982 &0.9396 &0.9018 &2.88\\
G-CASCADE~\cite{rahman2024g}&WACV 2024 &0.9096 &0.8465 &0.8872 &0.9489 & 0.8948 & 2.78 \\
\textbf{FANetv2 (Ours)} &- &\textbf{0.9186} &\textbf{0.8646} &\textbf{0.9058} &\textbf{0.9535} &\textbf{0.9096} &2.83 \\
\bottomrule
\end{tabular}
\label{tab:resultsbkai}
\end{table}

\begin{table}[t!]
\footnotesize
\centering
\caption{Result of models trained and tested on CVC-ClinicDB~\cite{bernal2015wm}.}
 \begin{tabular} {l|c|c|c|c|c|c}
\toprule
\textbf{Method} &\textbf{mDSC}  & \textbf{mIoU}  &\textbf{Recall}& \textbf{Precision} &\textbf{F2} &\textbf{HD} \\ 
\hline
U-Net~\cite{ronneberger2015u} &0.8973 &0.8440 &0.9105 &0.9118 &0.9036 &3.47\\
DeepLabV3+~\cite{chen2018encoder} &0.9339 &0.8882 &0.9351 &0.9368 &0.9341 &3.18\\
\textcolor{blue}{FANet}~\cite{tomar2022fanet} &\underline{0.8735} &\underline{0.8063} &\underline{0.9215} &\underline{0.8748} &\underline{0.8944} &\underline{3.76}\\
PraNet~\cite{fan2020pranet} &0.9419 &0.8925 &0.9469 &0.9406 &0.9444 &3.18\\
Polyp-PVT~\cite{dong2021polyp} &0.9193 &0.8635 &0.9342 &0.9232 &0.9271 &3.42\\
UACANet~\cite{kim2021uacanet} &0.9418 &0.8938 &0.9380 &\textbf{0.9517} &0.9389 &3.20\\ 
DuAT~\cite{tang2022duat} &0.9347 &0.8881 &0.9485 &0.9277 &0.9414 &3.23\\
CaraNet~\cite{lou2022caranet} &0.9251 &0.8748 &0.9324 &0.9202 &0.9292 &3.21\\
LDNet~\cite{zhang2022lesion} &0.9312 &0.8780 &0.9416 &0.9262 &0.9369 &3.33\\
TGANet~\cite{tomar2022tganet} &0.9434 &0.9015 &0.9430 &0.9461 &0.9429 &3.19\\
TransNetR~\cite{jha2024transnetr} &0.9107 &0.8628 &0.9044 &0.9380 &0.9065 &3.35\\
G-CASCADE~\cite{rahman2024g} &0.9374 &0.8910 &0.9483 &0.9349 &0.9416 &\textbf{3.18}\\
\textbf{FANetv2 (Ours)} &\textbf{0.9481} &\textbf{0.9039} &\textbf{0.9514} &0.9490 &\textbf{0.9496} &3.19\\
\bottomrule
\end{tabular}
\label{tab:results-cvc}
\vspace{-2mm}
\end{table}

\begin{figure*} [!t]
    \centering
    \includegraphics[width=0.65\textwidth]{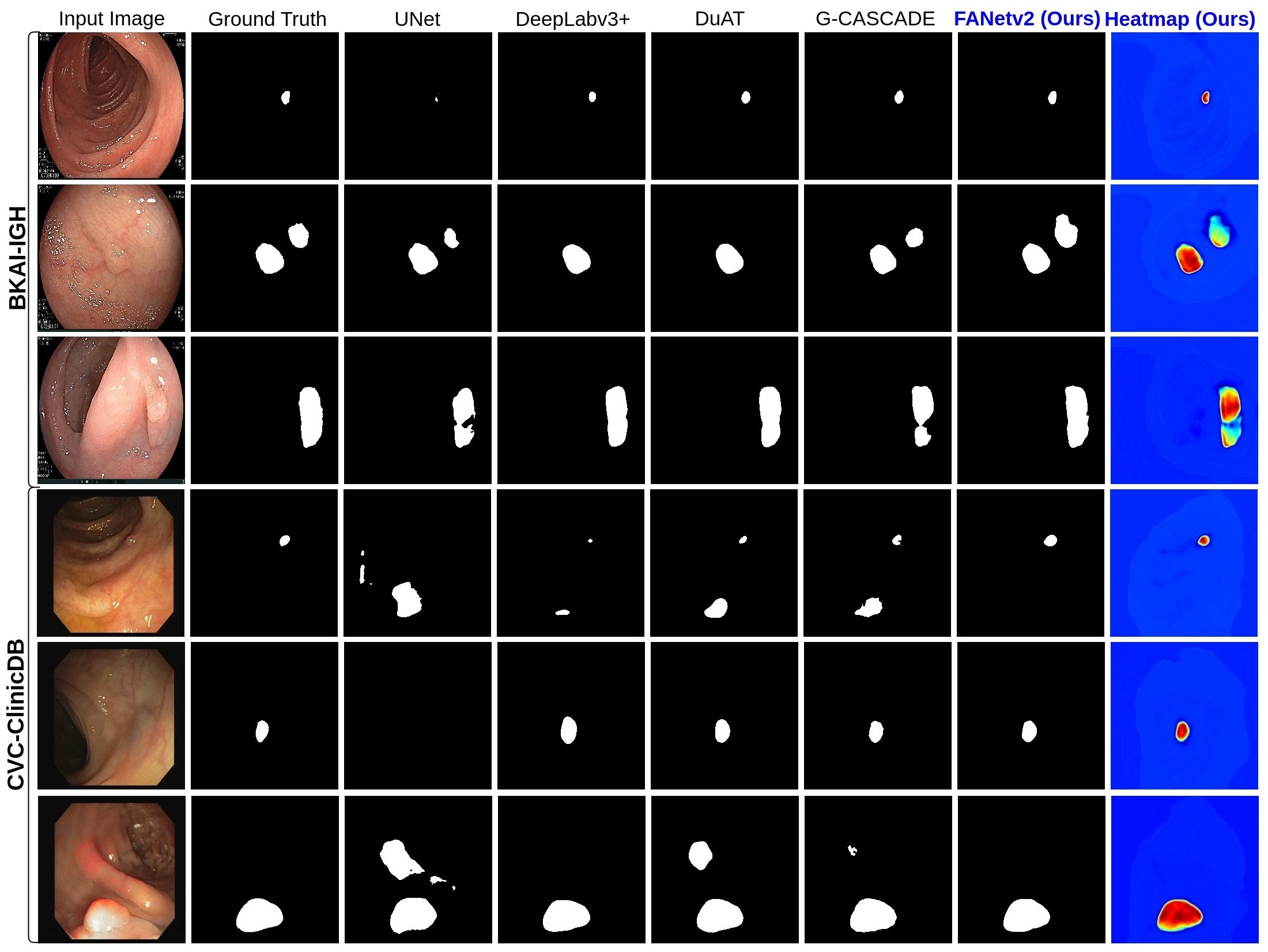}
    \caption{Qualitative results comparison on BKAI-IGH~\cite{lan2021neounet} and CVC-ClinicDB~\cite{bernal2015wm}. The heatmap shows the FANetv2's precision in accurately predicting polyps of various sizes and shapes.}
    \label{fig:qualitative1}
    \vspace{-5mm}
\end{figure*}

\subsection{Results}
\paragraph{\textbf{Results on BKAI-IGH dataset:}} Table~\ref{tab:resultsbkai} shows the results of all the models on BKAI-IGH dataset. The table demonstrates the superiority of FANetv2  with a high \ac{DSC} of 0.9186, \ac{mIoU} of 0.8646, a high recall of 0.9048, a precision of 0.9535,  low \ac{HD} of 2.83. It outperforms 12 \ac{SOTA} methods.  DuAT~\cite{tang2022duat}, TransNetR~\cite{jha2024transnetr} and G-CASCADE~\cite{rahman2024g} are the most competitive network to our FANetv2, where our network still outperforms DuAT and TransNetR by 0.46\%  and 0.79\% in \ac{DSC} respectively. These results suggest that FANetv2 is effective in segmenting polyps on different endoscopic imaging modalities such as WLI, BLI, FCI, and FICE.   

\paragraph{\textbf{Results on CVC-ClinicDB dataset:}} Table~\ref{tab:results-cvc} shows the results of all the models on the CVC-ClinicDB dataset. FANetv2 obtains a high \ac{DSC} of 0.9481, \ac{mIoU} of 0.9039, recall of 0.9514, precision of  0.9490, and low \ac{HD} of 3.19. The most competitive network to FANetv2 is TGANet~\cite{tomar2022tganet}.  Our model surpasses by TGANet by 0.47\% in \ac{DSC}, 0.24\% in \ac{mIoU} in metrics.


\begin{table*} [!t]
    
\centering
\caption{Comparison of model parameters, flops, and FPS for all models.}
\begin{tabular}{l|c|c|c|c|c} 
\toprule
\textbf{Method} &\shortstack{\textbf{Publication}\\ \textbf{Venue}} &\textbf{Backbone} & \shortstack{\textbf{Param.}\\ (\textbf{Million)}} &\shortstack{\textbf{Flops}\\\textbf{(GMac)}} & \textbf{FPS}\\ \midrule 

U-Net &MICCAI 2015 &-  &31.04	&54.75 &\textbf{94.99} \\
DeepLabv3+ &ECCV 2018 &ResNet50 &39.76 &43.31 &60.47\\
\textcolor{blue}{FANet} &TNNLS 2021 &- &\textbf{7.72} &94.75 &36.64 \\ 
PraNet &MICCAI 2021 &Res2Net50 &32.55 &6.93 &29.68\\ 
Polyp-PVT &ArXiv 2021 &PVTv2-B2 &25.11 &5.30 &47.54 \\
UACANet &ACMMM 2021 & Res2Net50 &69.16 &31.51 &29.09 \\ 
DuAT& Arxiv 2022 &PVTv2-B2 &24.97 &\textbf{5.24} &44.56\\
CaraNet & MIIP 2022 &Res2Net101 &46.64	&11.48 &20.67\\
LDNet& MICCAI 2022 &Res2Net50 &33.38 &33.14 &39.55\\
TGANet & MICCAI 2022 &ResNet50 &19.84	&41.88 &47.73\\
TransNetR &MIDL 2023 &ResNet50 &27.27 &10.58 &61.33 \\ 
G-CASCADE &WACV 2024 &PVTv2-B2 &26.63 &5.55 &44.75 \\ 
\textbf{FANetv2} & - &PVTv2-B3 &96.93 &21.59 &20.53 \\
\bottomrule
\end{tabular}
\label{algorithm_complexity}
\vspace{-2mm}
\end{table*}

\subsection{Comparison of FANet vs FANetv2}
FANetv2 has a 9.81\% improvement in BKAI-IGH datasets and a 7.46\% improvement in the CVC-clinicDB dataset. From Table~\ref{algorithm_complexity}, we can observe that FANetv2 has 96.93 million parameters, whereas FANet has only 7.72 million parameters. This is because FANet is built from scratch rather than using any pre-trained model as an encoder. However, FANetv2 uses a PVTv2 encoder. Regarding computational complexity, FANetv2 is more efficient as it has 21.59 GMac. FANet has an advantage in real-time processing with a processing speed of 36.64, whereas FANetv2 maintains a near-real-time processing speed of 20.53. To further show the significance of FANetv2, we provide animated .GIF files for the test results on polyp still frames and video sequences in the \textbf{Supp. material}. 

\vspace{-2mm}

\section{Conclusion}
We introduced FANetv2 architecture for automatic polyp segmentation. FANetv2 combines the Pyramid Vision Transformer (PVT) encoder with Feature Enhancement Dilated (FED) blocks alongside its unique mechanisms for integrating image and textual features through the auxiliary polyp attribute classification. The decoder's Mask Attention (MA) block significantly enhances its ability to accurately segment various polyp sizes and types. By integrating iterative feedback attention learning, text-guided attention, and refinement of prediction masks during testing, FANetv2 achieves the highest DSC and lowest HD values across two BKAI-IGH and CVC-ClinicDB datasets, outperforming FANet and other 11 SOTA methods. The qualitative results further validate the FANetv2's efficacy, thus highlighting its potential for clinical applications.


\bibliographystyle{splncs04}
\bibliography{ref}
\end{document}



\title{Transformer-Enhanced Iterative Feedback Mechanism for Polyp Segmentation}

\titlerunning{Transformer-Enhanced Iterative Feedback Mechanism for Polyp Segmentation}

  \maketitle      

\begin{table}[!htbp]
\footnotesize
\centering
\caption{Result of FANetv2 iteration trained and tested on BKAI-IGH dataset.} 
 \begin{tabular} {l|c|c|c|c|c|c}
\toprule
\textbf{Iteration} &\textbf{mDSC}  & \textbf{mIoU}  &\textbf{Recall}& \textbf{Precision} &\textbf{F2} &\textbf{HD} \\ 
\hline
1 &0.8840 &0.8139 &\textbf{0.9364} &0.8617 &0.9094 &2.85\\
2 &0.9181 &0.8635 &0.9081 &0.9499 &\textbf{0.9108} &\textbf{2.81}\\
3 &\textbf{0.9186} &\textbf{0.8646} &0.9058 &0.9535 &0.9096 &2.83\\
4 &0.9180 &0.8639 &0.9047 &0.9540 &0.9087 &2.84\\
5 &0.9177 &0.8636 &0.9043 &\textbf{0.9541} &0.9083 &2.84\\
\bottomrule
\end{tabular}
\label{tab:resultsbkai}
\end{table}

\begin{table}[!htbp]
\footnotesize
\centering
\caption{Result of FANetv2 iteration trained and tested on CVC-ClinicDB dataset.} 
 \begin{tabular} {l|c|c|c|c|c|c}
\toprule
\textbf{Iteration} &\textbf{mDSC}  & \textbf{mIoU}  &\textbf{Recall}& \textbf{Precision} &\textbf{F2} &\textbf{HD} \\ 
\hline
1 &0.8988 &0.8366 &\textbf{0.9568} &0.8700 &0.9266 &3.38\\
2 &\textbf{0.9481} &\textbf{0.9039} &0.9514 &0.9490 &\textbf{0.9496} &\textbf{3.19}\\
3 &0.9477 &0.9035 &0.9499 &0.9502 &0.9485 &3.20\\
4 &0.9475 &0.9033 &0.9494 &0.9504 &0.9482 &3.20\\
5 &0.9475 &0.9032 &0.9493 &\textbf{0.9505 }&0.9481 &\textbf{3.19}\\
\bottomrule
\end{tabular}
\label{tab:resultsbkai}
\end{table}

In addition to the results shown in the paper, we present the results of FANetv2's on still frames (BKAI-IGH dataset) and video sequence from the routine colonoscopy examination, along with the supplementary files.  The qualitative results on still frames and videos support the idea that FANetv2 can be useful in real-world clinical scenarios.